\documentclass{article}

\usepackage{times}
\usepackage{soul}
\usepackage{url}
\usepackage[hidelinks]{hyperref}
\usepackage[utf8]{inputenc}
\usepackage[small]{caption}
\usepackage{graphicx}
\usepackage{amsmath}
\usepackage{booktabs}
\usepackage{algorithm}
\usepackage{algorithmic}
\usepackage[dvipsnames]{xcolor}
\usepackage{multirow}
\usepackage{siunitx}
\usepackage{named}

\twocolumn 
\title{Watermark retrieval from 3D printed objects via synthetic data training}

\author{
\parbox{1.0\textwidth}
{\centering
Xin Zhang, Ning Jia and Ioannis Ivrissimtzis\\
Durham University\\
Department of Computer Science\\
Durham, DH1 3LE, UK\\
\{xin.zhang3, ning.jia, ioannis.ivrissimtzis\}@durham.ac.uk
}
}

\begin{document}	

\date{}
\maketitle
		
\begin{abstract}
We present a deep neural network based method for the retrieval of watermarks from images of 3D printed objects. To deal with the variability of all possible 3D printing and image acquisition settings we train the network with synthetic data. The main simulator parameters such as texture, illumination and camera position are dynamically randomized in non-realistic ways, forcing the neural network to learn the intrinsic features of the 3D printed watermarks. At the end of the pipeline, the watermark, in the form of a two-dimensional bit array, is retrieved through a series of simple image processing and statistical operations applied on the confidence map generated by the neural network. 
The results demonstrate that the inclusion of synthetic DR data in the training set increases the generalization power of the network, which performs better on images from previously unseen 3D printed objects. We conclude that in our application domain of information retrieval from 3D printed objects, where access to the exact CAD files of the printed objects can be assumed, one can use inexpensive synthetic data to enhance neural network training, reducing the need for the labour intensive process of creating large amounts of hand labelled real data or the need to generate photorealistic synthetic data.
\end{abstract}

\section{Introduction}
\label{section I}

In the past few years, deep Convolutional Neural Networks (CNNs) have transformed the field of computer vision, extending its domain of application into new directions. While CNNs are now the tool of choice in several fundamental computer vision tasks, such as image classification, semantic segmentation, image captioning, object detection, and human identification, nevertheless, their suitability for various specific applications still awaits to be verified, especially in situations where there is a shortage of annotated data. 

Watermark retrieval from 3D printed objects is one such application domain that remains less studied. {\em Watermarking}, a term used in a broad sense in this paper, is the embedding of information on a physical object or a digital file. It may serve various purposes, such as object or file authentication, copy control, protection against unauthorized alteration, or dissemination of machine readable information. The latter is an increasingly popular application, especially in the form of QR codes, and is the target application of the proposed method for watermarking 3D printed objects. In our study, similarly to QR codes, the watermark is a two-dimensional bit array, which is printed on a flat, non-slanted surface of the object in the form of small bumps arranged in a regular grid. 

The superior performance of CNNs over classical machine learning models lies in the tuning of millions of parameters stored in the deep hierarchical structures by training them with abundant data and the corresponding annotations. That means that the accurate retrieval of the bit array of the watermark from a digital image of a 3D printed object requires a large amount of training images, annotated with the locations of the watermark bumps. The standard training data acquisition process would require the 3D printing of several objects, which can be expensive, and hand-annotating images of these objects which in an error-prone and laborious process. 

In particular, the hand-annotation of this type of training data, as in \cite{zhang2018watermark}, has two known sources of error. Firstly, there is a human error introduced when the marked coordinates of the watermark bumps are not exactly on the centroid of the bump. Secondly, even if we assume semi-spherical bumps, as it is the case in our experiments, there is a systematic error introduced when a Gaussian kernel is applied on the marked coordinates to generate a smooth confidence map. Indeed, unless we manually adjust the size of the Gaussian kernel over each training image, which in practice it is infeasible, the training confidence maps do not adjust to the scale of the object's image, i.e., assuming objects of the same size, to the distance of the camera from the object. Moreover, in what is the most intrinsic source of error in that process, when the image of the object is taken from a low angle, the images of the bumps deviate considerably form the circular shape and will always be over or under-covered by the circular support of the Gaussians. 

To address the problems of human and systematic errors in the hand-labelling process, which confuse the CNN models and reduce their accuracy, we propose to use the Computer Aided Design models of the 3D printed models, which anyway exist since they are needed for 3D printing, to automatically produce synthetic training images with exact annotations. Apart from increasing the accuracy of the annotations, the proposed solution also reduces the cost of creating real training data, i.e. printing materials and labor, which means that we can have more variability in terms of object colors, illumination conditions and camera distance and angle.

Our approach follows recently proposed cost efficient solutions for bridging the gap between real and synthetic data using graphic generators such as UnrealStereo \cite{zhang2016unrealstereo}, which can generate photo-realistic synthetic data and the corresponding ground truths, and has been successfully applied in training neural networks for optical flow\cite{qiu2016unrealcv,barron1994performance}, semantic segmentation and stereo estimation. As UnrealStereo requires designers at artist level, domain randomization (DR)~\cite{tobin2017domain} has been recently proposed to address this limitation.  Domain randomization is currently used for robots to locate objects with simple shapes, forcing the network to focus on the essential features of the objects in the images rather than photorealism. 

\begin{figure}
	\centering
	\includegraphics[width=0.39\columnwidth]{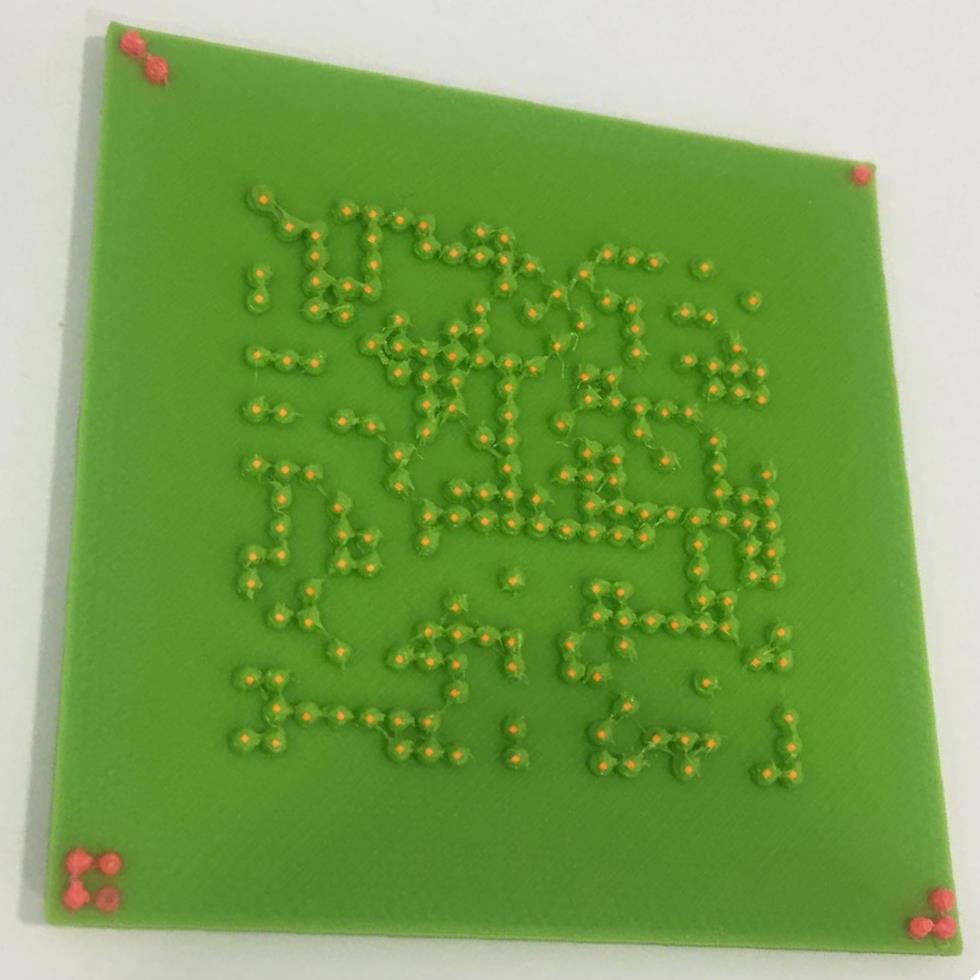} \hfill
	\includegraphics[width=0.58\columnwidth]{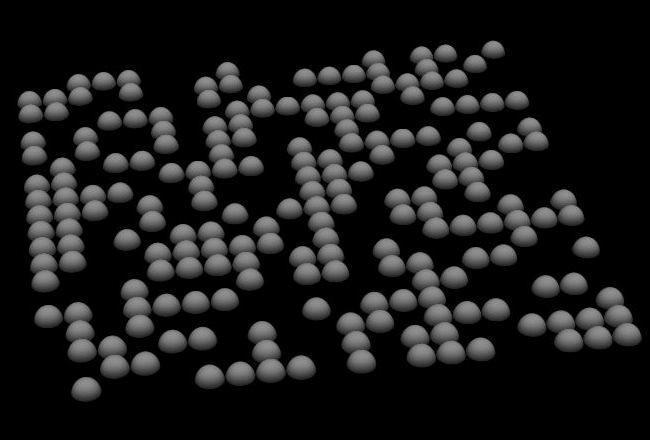} 	
	\caption{Real image with hand-annotated coordinates of the watermark centroids (left1). Ground thruth image generated by the simulator (left2).} 
	\label{confMap}
\end{figure}

\paragraph{Contribution.}  Addressing shortcomings in Zhang \textit{et al. }~\shortcite{zhang2018watermark}, which relies on a high cost, small, manually labelled real image dataset, in this paper we employ domain randomization (DR) for watermark retrieval from 3D printed objects, training the CNN with synthetic image data and confidence maps. The use of synthetic data not only reduces the cost of creating the training dataset, but also eliminates the human and the systematic errors of hand-labelling. The confidence maps are now precise in that they correspond exactly to the watermark bumps areas, see Figure~\ref{confMap}. The main contributions are summarized as follows:  
\begin{itemize}
	\item Application of the domain randomization method to generate synthetic data for a non-trivial problem, watermark retrieval from 3D printed objects. The practicability of our approach rests in the fact that no artistic input is required at any stage of the process. 
	\item Experimental investigation of the effect of the parameters of the synthetic data. 
\end{itemize}
Our experiments show that on images from previously unseen 3D printed objects the highest precision and recall rates are achieved when we train the neural network with a combined dataset consisting of synthetic DR and real data. A series of further experiments shows that the inclusion of the synthetic data in the training set increases the generalization power of the network, enabling it to learn the basic invariant features of the target objects. 

The rest of the paper is organized as follows. In Section \ref{section II} we discuss related work; 
in Section \ref{section III} we introduce our synthetic data image generator and our 
Convolutional Neural Network - 3D Watermarking (CNN-3DW); in section \ref{section IV} we test with real image 
data and explore how synthetic data support training; and finally we briefly conclude in 
section \ref{section V}.

\section{Related Work} 
\label{section II}

We review prior work in three domains: 3D model watermarking in section~\ref{subsection I}, object detection with CNNs in section~\ref{subsection  II}, and synthetic data with domain randomization in section~\ref{subsection III}.

\subsection{Digital 3D Watermarking} 
\label{subsection I}

The focus of the research on 3D model watermarking is digital 3D model watermarking \cite{ohbuchi2002frequency,bors2006watermarking,luo2011surface} aiming at the development of algorithms that are robust against a variety of malicious or unintentional attacks, including 3D printing and rescanning. The watermark survivability under such an attack has been tested in \cite{macq2015applicability}. Surface watermarking algorithms resilient to the attack have been proposed in \cite{hou20153d,Hou17} the former encoding the watermark as modulations of horizontal slices of the 3D model. A survey of digital rights management and protection technologies for 3D printing can be found in \cite{Hou18}. 

The above 3D printing related approaches have a different objective than ours, that is, they want to protect the 3D digital model rather than embed a physical object with machine readable information. As a result, the retrieval process requires the use of a laser scanner for the reconstruction of the original 3D model, while we apply instead standard computer vision techniques and never need to reconstruct a digital 3D model from the 3D printed object. In practice, their different objective also means that the capacity of the proposed algorithms is usually low, just enough for the purposes of model authentication.

\subsection{Object Detection with CNNs} 
\label{subsection  II}

Convolutional neural networks, AlexNet \cite{krizhevsky2012imagenet}, VGGNet \cite{simonyan2014very}and GoogleNet \cite{szegedy2015going} being some famous examples, have demonstrated impressive performance on object detection tasks and are considered particularly well suited for feature extraction and classification. More recently, CNN models  have been used in crowd counting tasks~\cite{zhang2016single,mundhenk2016large}, often formulated as regression problems with the raw images as input and a feature map characterizing the density of the crowd in the image as output. 

Here, we take the simple but effective approach of training a CNN with synthetic image data for confidence map extraction from real images. 
The choice to use a CNN was informed by an earlier approach to the problem where we used local binary patterns (LBPs)~\cite{ojala2000gray}. However, LBPs did not perform very well under adverse conditions, particularly when background patterns were too prominent; under extreme uneven illumination; or under unfavorable camera viewpoints. 
\subsection{ Synthetic dataset creation and domain randomization} 
\label{subsection III}

The synthetic data methods support accurate ground truth annotations, and are cheap alternatives to annotating images. They have been widely used in deep learning. Jaderberg \textit{et al.}~\shortcite{jaderberg2014synthetic} and Gupta \textit{et al.}~\shortcite{gupta2016synthetic} paste real text images to blank backgrounds and real natural background respectively. Dosovitskiy \textit{et al.}~\shortcite{dosovitskiy2015flownet} combines renderings of 3D chairs images with natural image backgrounds, such as city, landscape and mountains, to train FlowNet. Handa\textit{et al.}~\shortcite{handa2016understanding} build indoor scenes using CAD models to understand real world indoor scenes. Atapour-Abarghouei and Breckon~\shortcite{atapour2018real} train a depth estimation model using synthetic game scene data. Zhang \textit{et al.}~\shortcite{zhang2016unrealstereo} develop a synthetic image generation tool to analyse stereo vision. The first three of the above methods use existing datasets and models, while the last two require artist level synthetic image data. 

Domain randomization is a simpler technique for training models on simulated images. It randomizes some of the parameters of the simulator and generates a dataset of sufficient variety consisting of non-artistic synthetic data~\cite{tobin2017domain,tremblay2018training}. Here, we use domain randomization on our own 3D watermarked models, as there are no such datasets avaliabe, and train a CNN to retrieve watermarks from images of the corresponding 3D printed objects.   
\section{Method} 
\label{section III}

In this section we describe the generation of the synthetic training dataset and present the proposed framework for automatic 3D watermark retrieval.

% *************************************************

\subsection{Synthetic Image Generator}

To prevent the model from overfitting to unwanted features such as object colour, surface texture, bump size and shape, and camera view-point, the following aspects were considered when creating the training dataset:

\paragraph{The 3D shapes.} To simulate the expected diversity of 3D printed objects in a practical scenario, the generated 3D models should have a wide range of colors and textures, and be embossed with bumps of various sizes and shapes. Eventually, any type of real data would appear as variation of some of the synthetic data, and their bumps will be identified correctly regardless of their peculiarities. 

\paragraph{Camera view-point.} Introducing training images captured from arbitrary view angles increases the robustness of the model, thus we rotate the virtual camera of the simulator around the model. 

\paragraph{Scene illumination.} In a practical scenario, when we capture images of 3D printed objects we will not be able to control the illumination conditions. Thus we use multiple light sources, eliminating the influence of light intensity and shadow. 

\paragraph{Scene background.} By padding the captured images onto random backgrounds we reduce the sensitivity to background noise. The background photos are sampled from available public datasets of indoor and outdoor scenes.

\paragraph{Software.} For the synthetic data generation task we adopted Pov-Ray \cite{povray}, since it is open-source and supports several geometric primitives for constructive solid geometry, which is a popular modelling method for designing objects for 3D printing. 

\begin{figure}
	\centering
	\includegraphics[width= 0.8\columnwidth, height=3cm]{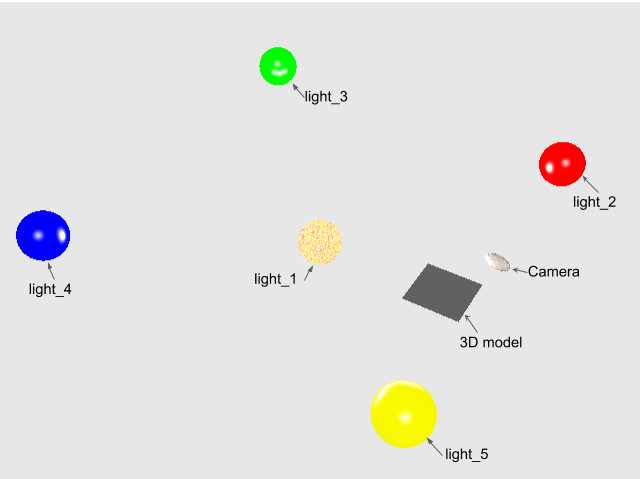} 
	\caption{Synthetic image generator.}
	\label{fig:pov ray model}
\end{figure}

The setup of the synthetic image generator is demonstrated in Figure \ref{fig:pov ray model}. It consists of five light sources, one 3D model, and one camera. When generating data, the camera revolves around the synthetic model to capture images under various angles; simultaneously, the synthetic model orbits light\_1, i.e. the primary illuminating source placed in the middle, while rotating around its own axis, to simulate variant beam directions. A random position shift of the camera within a moderate range is introduced to further increase the randomness of view points and scale, while additional illuminating sources lights\_2-5 are introduced for simulating a complex illumination scenario. As to the 3D model itself, we introduce color randomness and a wide range of optional textures to simulate the appearance of 3D printed objects. Note that camera's orbital speed is slightly faster than the angular velocity of the rotating model. 

We set Pov-Ray's float identifier clock at 120 frames per cycle with the initial frame at clock value 0.0 and the last frame at value 1.0.  

\paragraph{Textures and colors.}  Model textures are randomly picked from Pov-Ray's files at various scales. The RGB colors are randomized at \(\alpha*clock, \beta*clock, \gamma*clock\), where \( \alpha, \beta, \gamma \) $\in \left(0, 5 \right)$ are random numbers. 

\paragraph{Scene lighting.} To approximate a realistic environment, we use five different light sources. The centre (light\_1) is a parallel light source and around it in four different orientations we placed: a near point light (light\_2), a conical spotlight (light\_3), an area light (light\_4) and a fading light (light\_5). 

\paragraph{Pose and Camera.} Within each clock cycle, the synthetic 3D model rotates around its own axis by \(360*clock*30\) degrees and revolves around the central light by \(360*clock\) degrees. The camera revolves by \(360*clock*12\) degrees and its movement along the vertical axis is set to \(jStart+jHeight*(1-\cos(4\pi(clock-jStart)))/2\). The synthetic image is captured at pixel size \(3072*4096\).

\begin{figure*}
	\centering
	\includegraphics[width=0.24\linewidth]{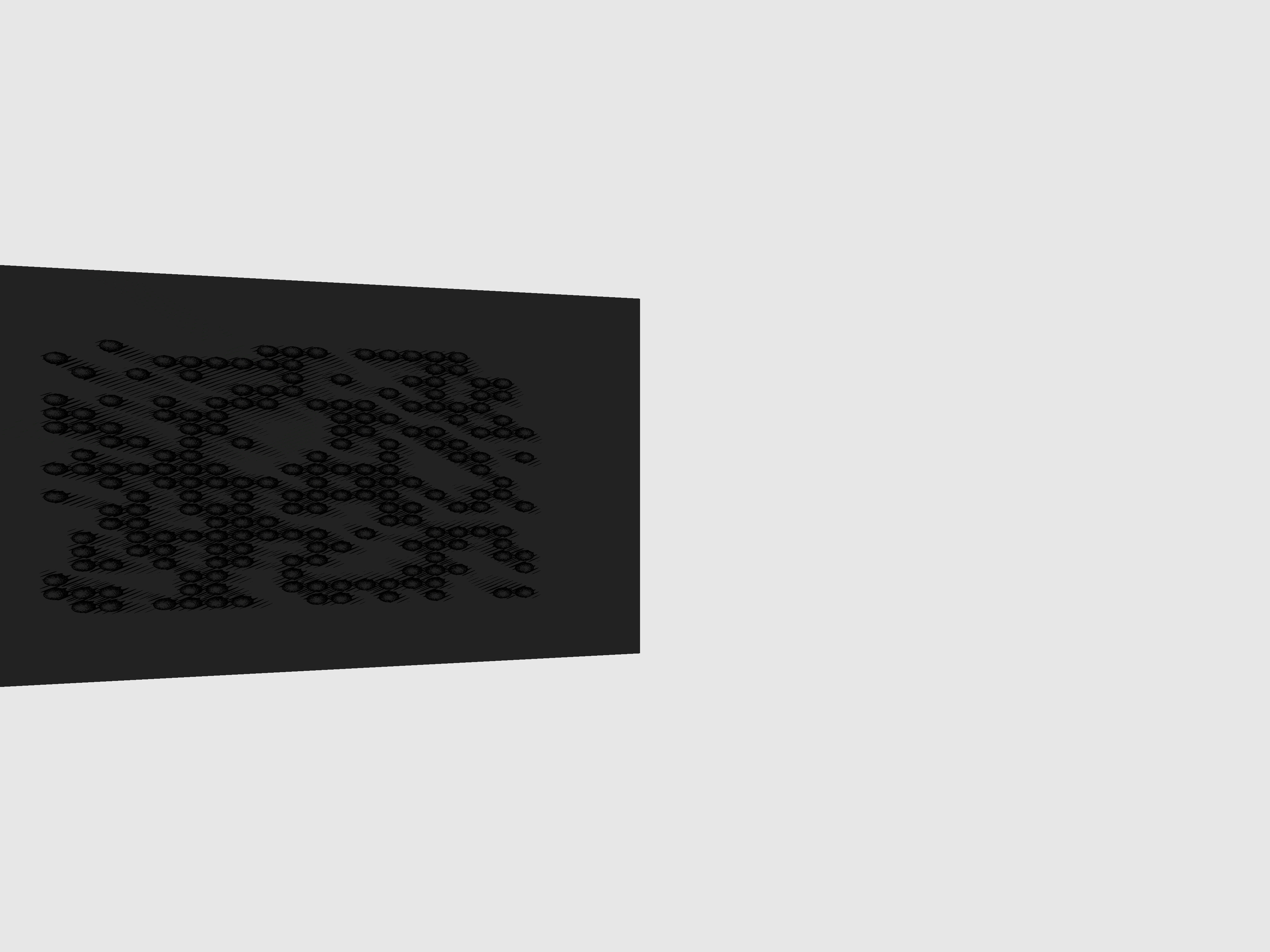} \hfill 
	\includegraphics[width=0.24\linewidth]{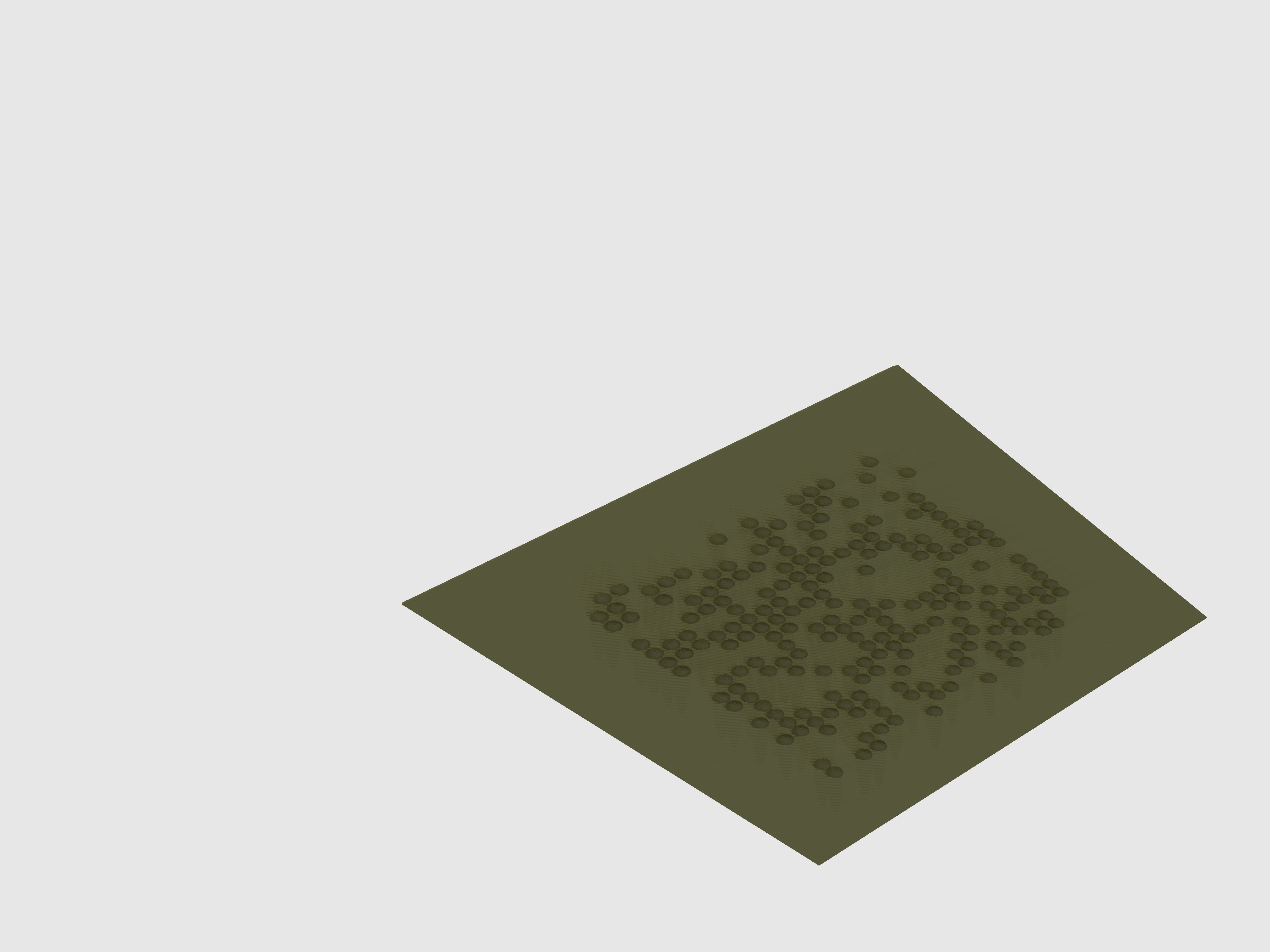} \hfill 
	\includegraphics[width=0.24\linewidth]{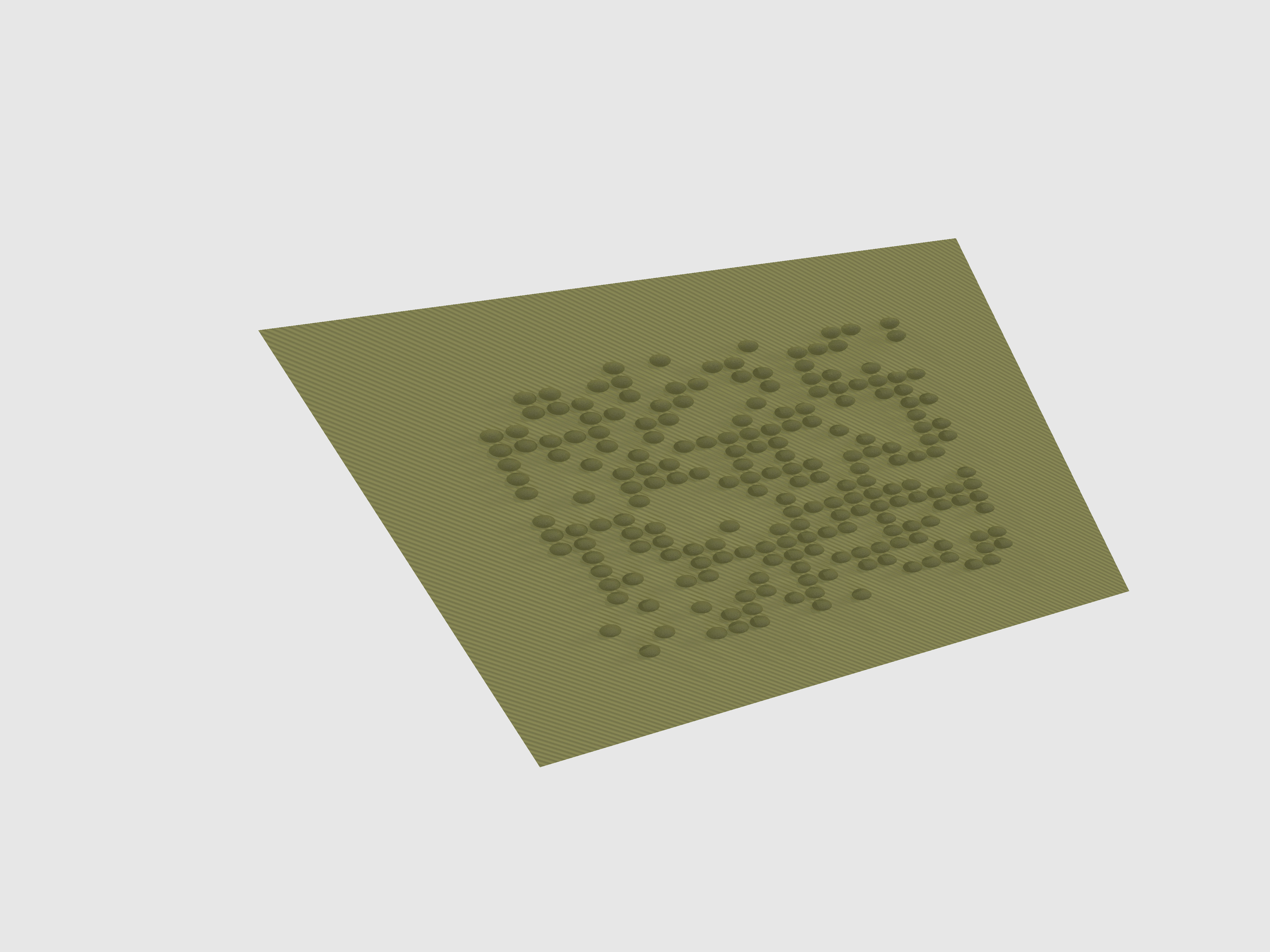} \hfill 
	\includegraphics[width=0.24\linewidth]{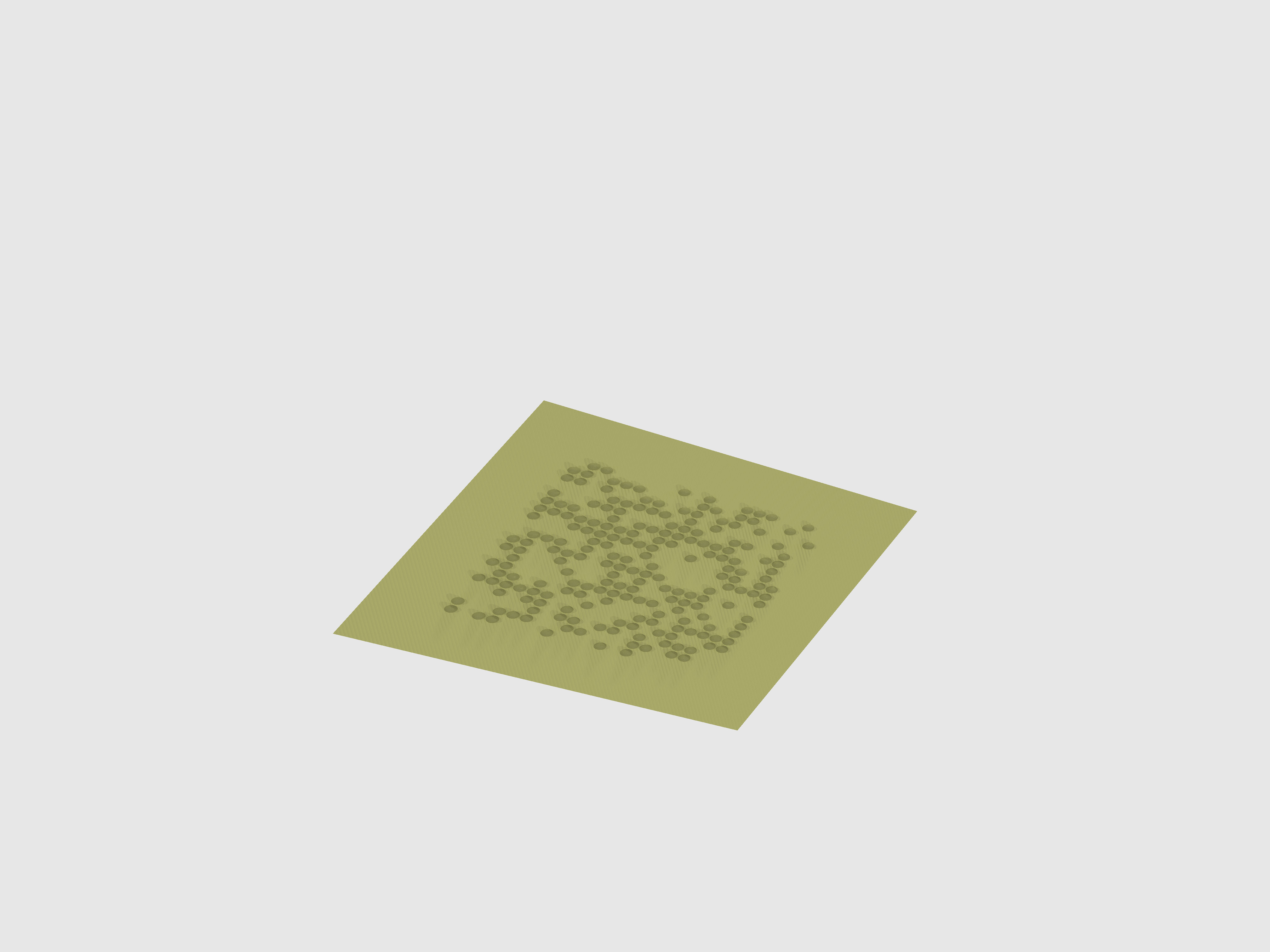} 
	\includegraphics[width=0.24\linewidth]{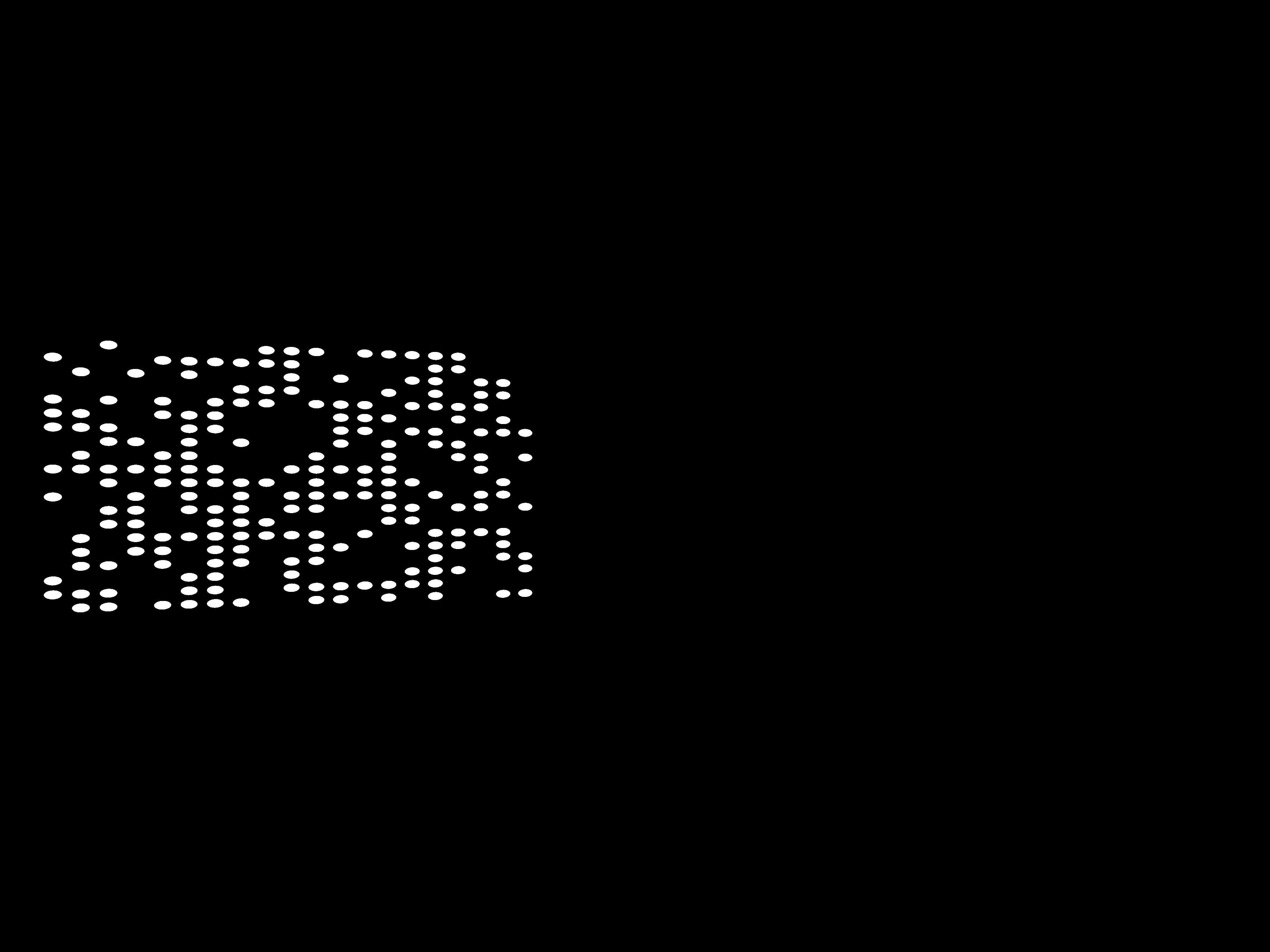} \hfill 
	\includegraphics[width=0.24\linewidth]{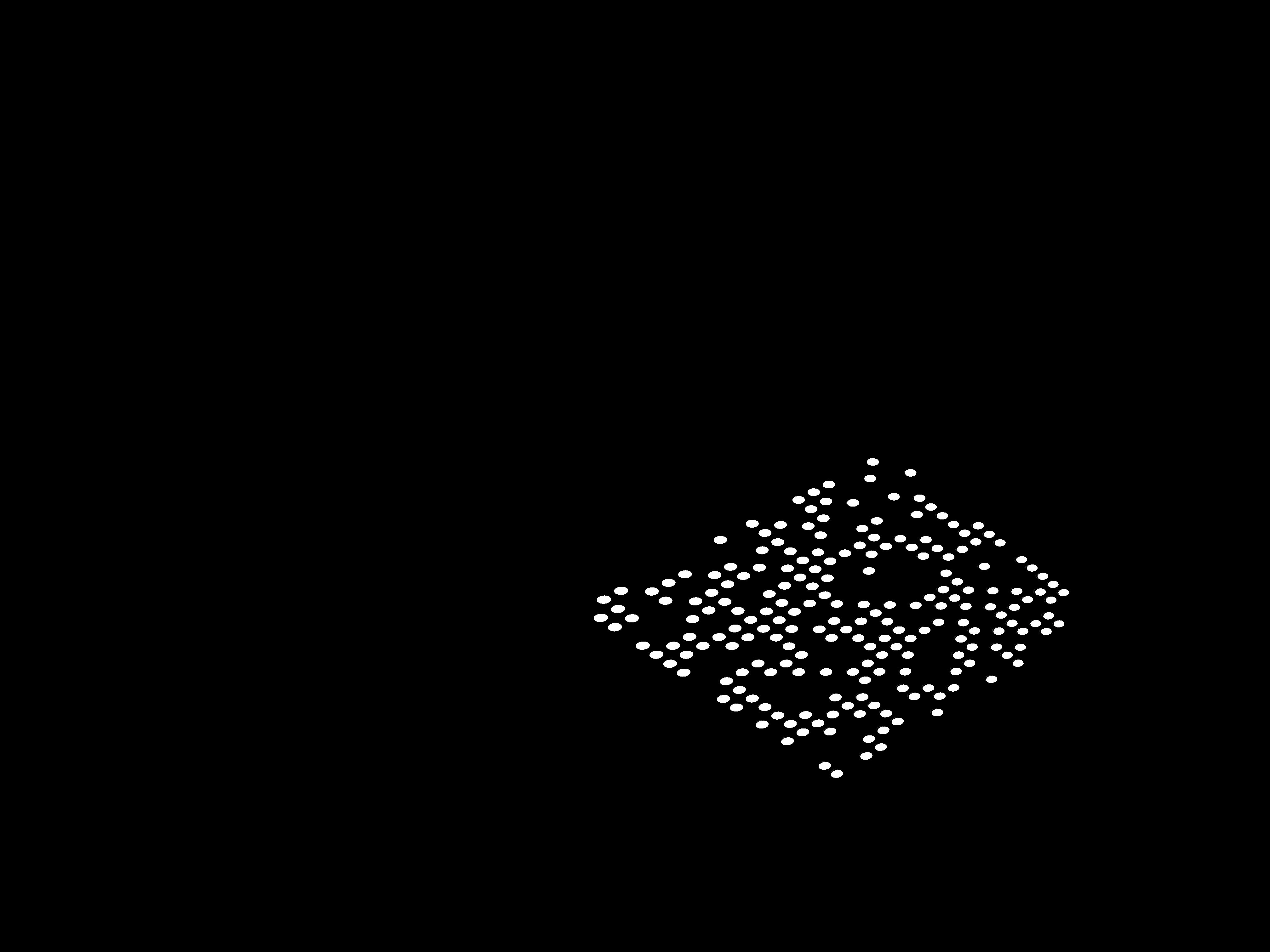} \hfill 
	\includegraphics[width=0.24\linewidth]{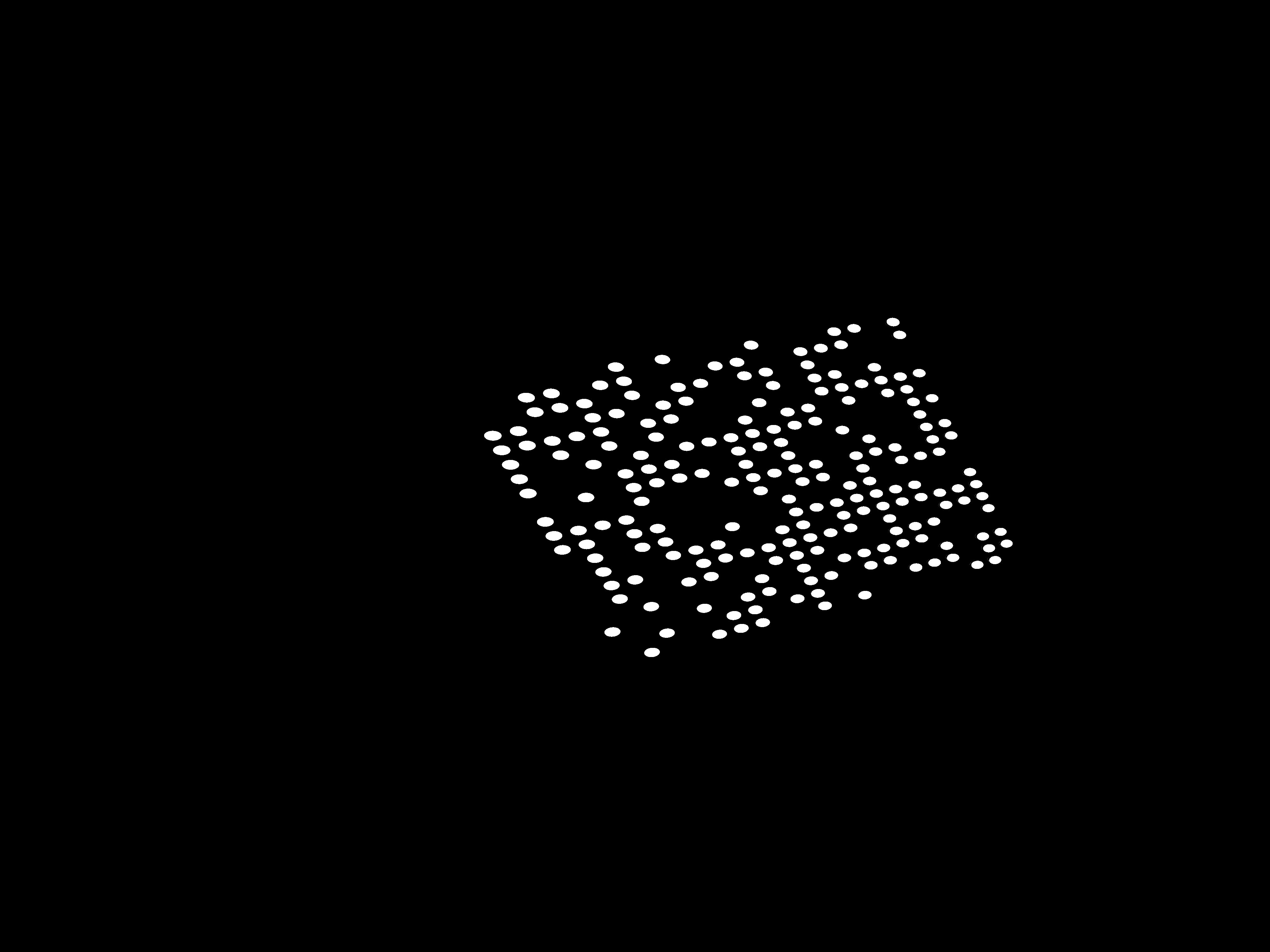} \hfill 
	\includegraphics[width=0.24\linewidth]{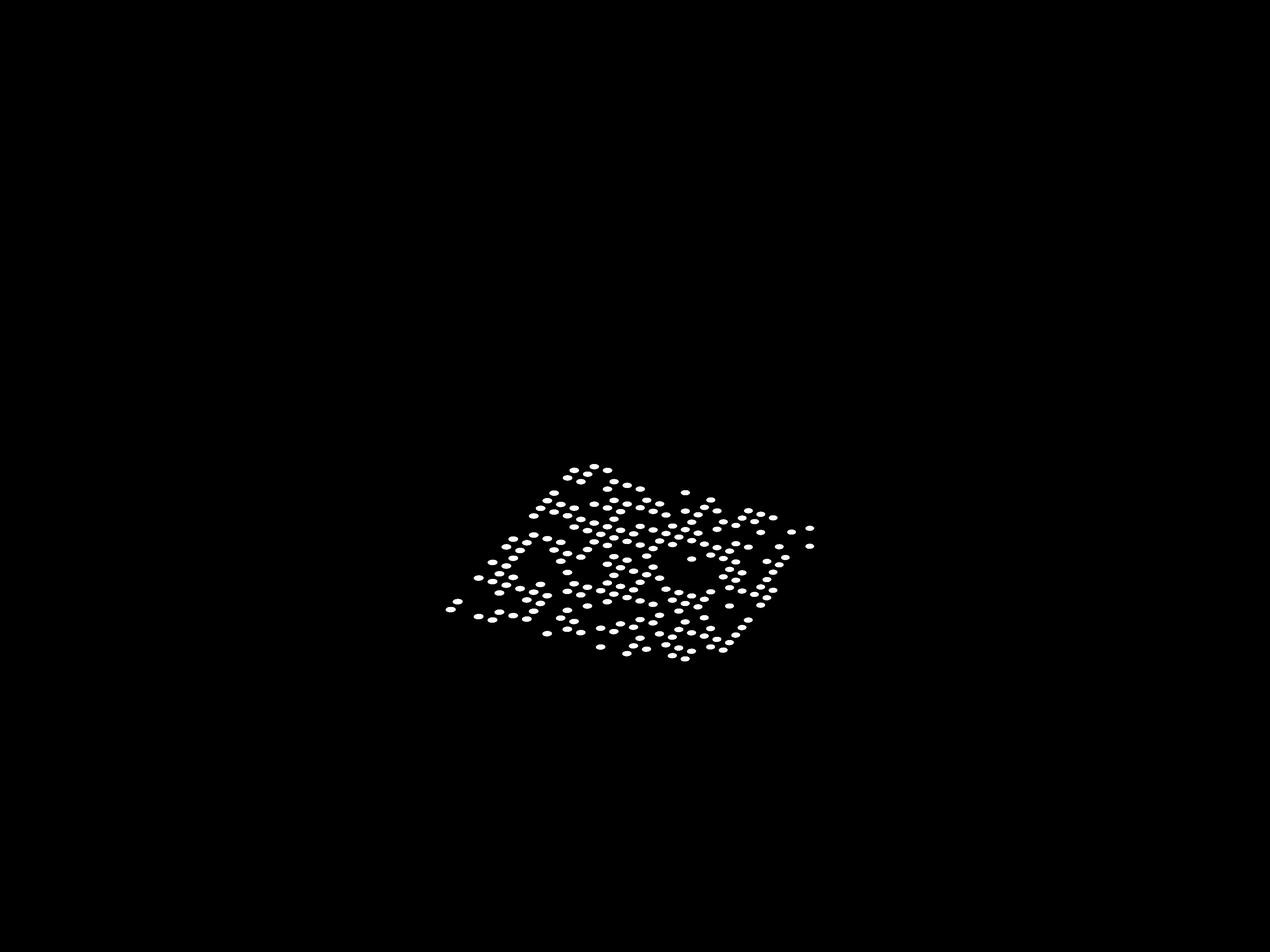} 
	\caption{Synthetic image and corresponding ground truth at different clock values.}
	\label{fig: data and ground truth}
\end{figure*}

The corresponding annotations are generated simultaneously with the synthetic image, as shown in Figure~\ref{fig: data and ground truth}. For generating annotations, we keep  the same the camera pose and light parameters and change the watermark bumps to full white while the rest of the model is set to black. The produced synthetic annotation is an exact map of the bump regions.

% ********************************************************* 

\subsection{Watermark detector}
Retrieving watermark information from 3D bumps is a dot detection problem. We tackle it by combining a deep convolutional neural network named \textbf{CNN-3DW} and an image registration module which processes the output of \textbf{CNN-3DW} to retrieve the watermark bit matrix.

% ********************************************************** 

\subsubsection{CNN-3DW}
\begin{figure*}
	\centering
	\includegraphics[width=\linewidth]{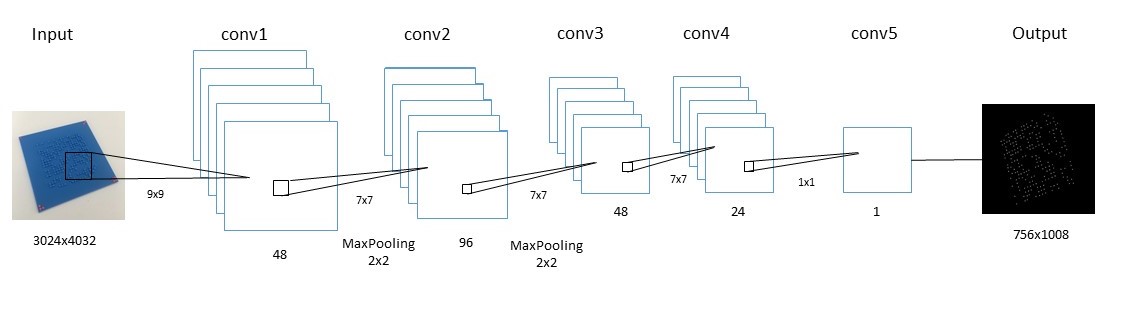}
	\caption{The architecture of the proposed CNN-3WD.}
	\label{fig:CNN-3WD}
\end{figure*}

The CNN-3DW is based on the network model shown in Figure \ref{fig:CNN-3WD}. It consists of five convolutional layers with kernel sizes \(9*9, 7*7, 7*7, 7*7, 1*1\) and feature map channels 48, 96, 48, 24, 1, respectively. The Rectified Linear unit (ReLu) is used as activation function. Each of the first two layers is followed by a standard max pooling layer, increasing the receptive field. For all the convolutional layers, we set the step size as one and do zero padding so that the image size is not altered by the convolutional operations. Since the model is fully convolutional, it can handle different input sizes. The reason we designed our network based on FCN rather than a deeper architecture such as VGG and ResNet is that we found in our experiments that the output of FCN was more stable. Indeed, locating dots on a single colored plate is a relatively simple task and the adoption of deeper models is more resource expensive. Further, as shown in Figure \ref{fig: data and ground truth}, the ground truth image composes of mostly zero value pixels indicating the irrelevant background, and by stacking non-linearity active functions such as ReLU can easily cause the network model to predict all-zero.  

Because our training data comes from inexpensive synthetic images with controllable parameters, data augmentation such as mirroring, translational shift, rotation and relighting is not necessary during the training phase. Instead, for time and resource efficiency, since the original images are oversized for graphics cards with small capacity, we randomly cropped each image to smaller patches of uniform size (1024*1024).

We implemented CNN-3DW in Pytorch and adopted the mean squared error (MSE) for computing loss. The model is optimised using Adam \cite{kingma2014adam}, the initial learning rate is \(1e -4\) and is decreased to \(1e-5\) after 150 epochs. We used a single NVIDIA GeForce GTX TITAN X graphics card to train our network with a batch size of 12.

% **********************************************************

\subsubsection{Image registration, processing and matrix retrieval}

\begin{figure*}
	\centering
	\includegraphics[width=0.24\linewidth]{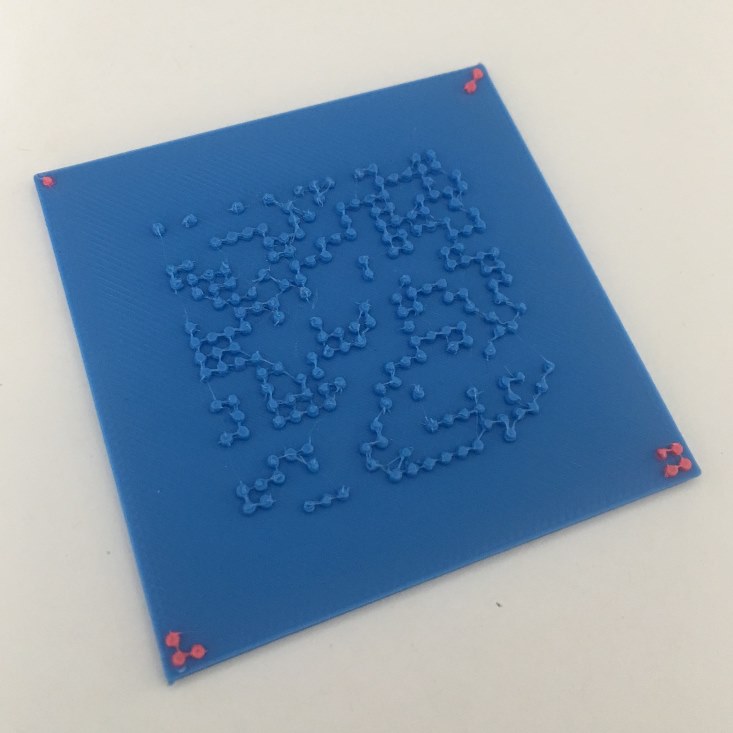} \hfill 
	\includegraphics[width=0.24\linewidth]{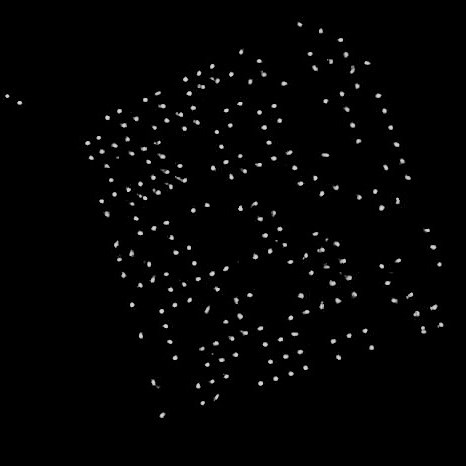} \hfill 
	\includegraphics[width=0.24\linewidth]{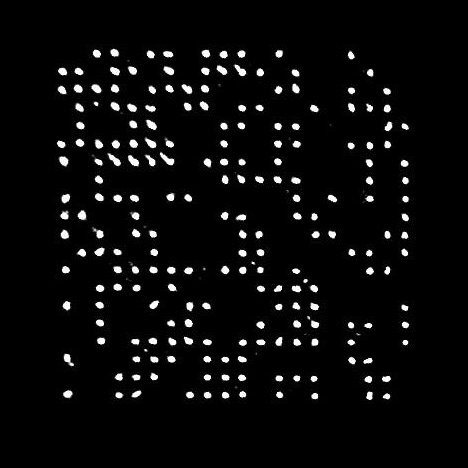} \hfill 
	\includegraphics[width=0.24\linewidth]{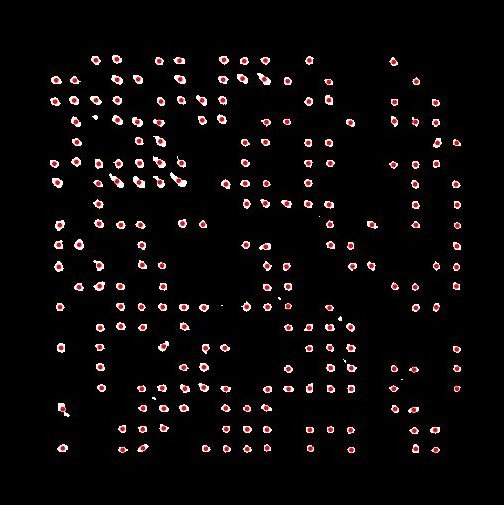} 
	\caption{The test image (left1); the confidence map outputted by CNN-3DW (left2); the binary confidence map obtained after regularisation followed by thresholding (left3); the detected centroids of bit value 1 regions (left4)}
	\label{fig:seventh}
\end{figure*}

Using the trained CNN-3DW model we are able to estimate a confidence map $\mathbf{Y}$ of the watermark bumps in the image $\mathbf{X}$. To extract the embedded information matrix from the estimated confidence map $\mathbf{Y}$ is still non-trivial. Indeed, due to variability in the illumination conditions and surface textures inadvertently introduced by the popular, inexpensive printers based on Fused Deposition Modelling (FDM) technology, a large amount of background noise can be introduced, posing great challenges to the automatic watermark region localization. Since a significant bias in the watermark region localization would lead to catastrophic error in the following stages, we decided to tackle this issue by printing four \textit{landmarks} at the corners of the watermark with distinguishable colours (see Figure \ref{fig:seventh}). During retrieval, the watermark regions are easily located by finding the differently colored \textit{landmarks} at the four corners. 

Further, since we do not require from the user to align their camera with the 3D information matrix when capturing images, the generated confidence map needs to be normalized before retrieving the information. We use image registration \cite{brown1992survey} to transform the quadrilateral watermark region in the confidence map $\mathbf{Y}$ to a square region (see Figure \ref{fig:seventh}), using Matlab's built-in tools to map the \textit{landmarks} to the four corners of a square region. We denote the normalized watermark region of the confidence map as $\mathbf{Y}^r$ and process it to extract the embedded information matrix $\mathbf{M}$. 

The registered confidence map $\mathbf{Y}^r$ is first binarized using a threshold value $t$,
\begin{equation}
\label{equ:binarize}
\hat{\mathbf{Y}}^r_{ij} =
\begin{cases}
1 \quad \text{if $\mathbf{Y}^r_{ij} > t$ ,}
\\
0 \quad \text{otherwise}.
\end{cases}
\end{equation}
where $t =\beta\times Thre$, $Thre$ is the Otsu threshold \cite{otsu1979threshold} of $\mathbf{Y}^r$ and $\beta$ a user defined parameter. In our experiments, we empirically set $\beta=0.35$, which seems to optimize performance. The binary confidence map $\hat{\mathbf{Y}}^r$ is visualized in Figure \ref{fig:seventh}~(left3). Next, we call Matlab's {\em regionprops} function to detect the connected regions of the binary image and obtain estimates of their centroids and their two semi-axes. If the the sum of the two semi-axes is above a threshold, a bit value 1 will be assigned to the centroid of that region as shown in Figure~\ref{fig:seventh}. 

As a final step, we need to extract the information matrix $M$ from the coordinates of the regionprops centroids that have been assigned bit values 1. Ideally, there would exist only $m$ (the number of rows and columns of $\mathbf{M}$) unique values for the $x$ and $y$ coordinates, however, due to imperfection in the previous steps, the coordinates have biases. Instead, we use $K$-means clustering on the $x$ and the $y$ coordinates of the centroids, respectively. We set the number of clusters as $m$ and rank the $m$ cluster centers. The point falling into the $i$-th and $j$-th clusters, respectively, will correspond to the $(i,j)$ entry of the matrix $\mathbf{M}$.

\section{Experiment} 
\label{section IV}

To assess the effectiveness of training the neural network model with synthetic data, we conducted a series of experiments and analyzed the results, evaluating the performance of proposed technique in various situations. 

% *************************************************

\subsection{Training datasets: 3DW-real and 3DW-syn}

For our 3DW-syn dataset, we generated 2K images each one carrying a randomly generated  $20 \times 20$ watermark. As described in the section~\ref{section III}, the model of the watermarked object is assigned a random color and texture, scenes are rendered under random camera poses and illuminated from light sources of various types. For comparison, we used a hand-labelled dataset (3DW-real) of 255 images of watermarked 3D printed objects \cite{zhang2018watermark}. Although we use more synthetic than real images, the cost of creating the 3DW-syn dataset was lower, and moreover 3DW-real was augmented in more ways than 3DW-syn. In particular, we applied random crop, brightness and contrast setting to both datasets, while 3DW-real was also augmented with random flips and random resizing. In all cases, training was stopped when the validation error was stable and the best weights were saved. 

For testing, we used 35 images (Test\_A) which are taken from same 3D printed models used in 3DW-real, but with different illumination conditions and camera poses, and 20 images (Test\_B) taken from other 3D printed models made from materials that were not used in the 3DW-real dataset. Detection performance was evaluated using recall TPR=TP/(TP+FN) and precision PPV=TP/(TP+FP), where in our case TP and FN are the numbers of value 1 bits retrieved correctly and correspondingly incorrectly, while FP is the number of value 0 bits retrieved incorrectly. 

% *************************************************

\subsection{Experimental Setup}

We evaluate the watermark retrieval capabilities of the proposed technique under the use of various training datasets, and provide an ablation study to further explore the effectiveness of using synthetic data.

\subsubsection{Watermark retrieval}

We evaluate the performance of the proposed framework on three scenarios: neural network model trained with real data (3DW-real) only, synthetic data (3DW-syn) only, and the combined dataset (3DW-real + 3DW-syn). Precision and recall are tested separately on Test\_A and Test\_B and the results are shown in Table~\ref{table:t1}. Note that Test\_A consists of images from the same 3D objects as the images of the (3DW-real) training dataset, while Test\_B contains images from different objects. 

\begin{figure*}
	\centering
	\includegraphics[width=\linewidth]{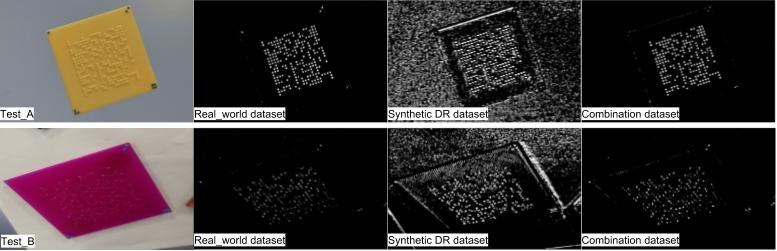}
	\caption{Confidence maps of images in Test\_A and Test\_B using CNN-3DW trained on: real-world dataset (left2), synthetic dataset (left3) and their combination (left4).}
	\label{fig: experiment1}
\end{figure*}

Figure \ref*{fig: experiment1} shows indicative outputs of the CNN-3DW on the Test\_A and Test\_B dataset, after being trained on either of the three training datasets we used. Notice that even though our network has never seen a real image when trained on (3DW-syn) only, it is able to successfully detect most watermark bumps. Moreover, when the network was trained on the combined (3DW-real + 3DW-syn), on the most challenging Test\_B it outperformed the network trained on (3DW-real) by a 15\% in the recall value and 5\% in precision. This surprising result illustrates the power of such a simple technique for bridging the reality gap, and rectifying the human and systematic error disadvantage in manual labeling.

\begin{figure}
	\centering
	\includegraphics[width=\columnwidth]{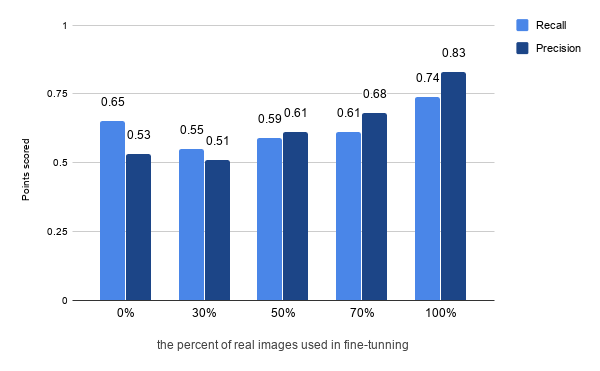}
	\caption{Percent of real images used in fine-tuning.}
	\label{fig: fine-tuning}
\end{figure}

Next, we study the effect of fine-tuning~\cite{yosinski2014transferable} the number of real images we add to (3DW-syn) to create a combined training dataset. For fine-tuning, the gradient was allowed to fully flow from end-to-end, the other parameters kept unchanged and the CNN-3DW was trained until convergence. Results with varying percentages of (3DW-real) added to the training dataset are shown in Fig~\ref{fig: fine-tuning}. We note that when a combined dataset is used, recall and precision increase with the number of real-images included. However, when less than 70\% of (3DW-real) is included the recall is lower compared to training with (3DW-syn) only. We hypothesize that when (3DW-syn) is mixed with few real images only, the real images confuse the neural network, preventing the learning of features from (3DW-syn).
\begin{table}
	\centering
	\resizebox{1\columnwidth}{!}{
	\begin{tabular}{lcccc}
		\toprule
		\multirow{2}{*}{Dataset} &
		\multicolumn{2}{c}{Test\_A} &
		\multicolumn{2}{c}{Test\_B}
		 \\
		& {Recall} & {Precision} & {Recall} & {Precision}\\
		\midrule
		Real-world dataset &\bfseries 0.85 & \bfseries0.89 & 0.57 & 0.75  \\
		Synthetic dataset & 0.65 & 0.53 & 0.62  & 0.55\\
		Combination & 0.74 & 0.83 & \bfseries0.72 & \bfseries0.80 \\
		\bottomrule
	\end{tabular}
}
\caption{Performance comparison among training datasets: real-world dataset, synthetic dataset and their combination, on Test\_A and Test\_B.}
\label{table:t1}
\end{table}

\subsubsection{Ablation study}

\begin{figure}
	\centering
	\includegraphics[width=\columnwidth]{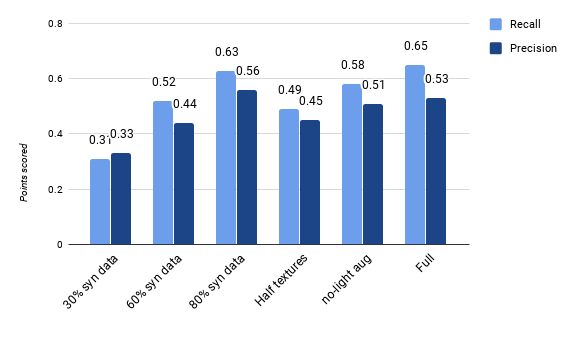}
	\caption{The effect of controlling individual components of the DR data generation procedure.}
	\label{fig: ablation}
\end{figure}

To study the effect of individual DR parameters we conducted an ablation study. In all cases we train with synthetic data only and since no real images are used we can safely use the larger Test\_A for testing. 
% by systematically controlling one of them once without pretrain. 
Figure~\ref{fig: ablation} shows the results of controlling individual components of the DR data generation procedure as described in more detail below.

\paragraph{Training dataset size.} In this experiment, we train with a percentage only of (3DW-syn), keeping all other parameters unchanged, and we study the effect of the training dataset size upon performance. The results show that recall and precision increase with the size of the dataset, but after a certain level recall and precision rise little.

\paragraph{Texture.} In this experiment, keeping other parameters unchanged, we create two sets of 1.5K synthetic images each, one utilizing half only of the available texture types, the other utilizing the full set of available textures. When the textures types applied on the watermark bumps were limited, recall and precision dropped to 0.49 and 0.45. 

\paragraph{Data augmentation.} In this experiment, during training with the full (3DW-syn), we turned off image brightness and contrast data augmentation. The recall falls to 0.58 while the precision barely drops by 0.02. Note that, in this case, the effect of lighting adjustments on already generated images is smaller than the effect of a changing a simulator parameter such as the number of texture types available. 

\subsubsection{Training strategies}

Finally, we studied the effect of pretrained weights upon the performance of CNN-3DW. In this experiment, we initialized the weights of the CNN-3DW network using random weights, the weights of the (3DW-real) trained network, and the weights of the (3DW-syn) trained network, respectively. In all cases, we retrained on (3DW-real + 3DW-syn) and tested on Test\_B. The results are shown in Table~\ref{table:t2}. 

First, note that the (3DW-real) weights achieve recall and precision of only 0.58 and 0.69 respectively, showing that the real-image weights are not very useful for training a network to operate on the Test\_B dataset. This is, in fact, a significant challenge with networks today, namely, that they often fail to transfer from one dataset to another. In contrast, the (3DW-syn) weights achieve better recall and precision rates, indicating that synthetic data enable the network to learn features which are invariant to the specific dataset. Finally, the best rates are achieved with random weights, i.e. when synthetic DR data are mixed with real-image data and train the network without pretrain. Indeed, training with synthetic data only produces inferior results, while training with real data only does cope well with variabiliy in parameters such as printing materials. 

\begin{table}
	\centering
	\resizebox{1\columnwidth}{!}{
		\begin{tabular}{cccc}
			\toprule
			 & Random weights & (3DW-real) weights & (3DW-syn) weights	\\
			\midrule
			Recall &\bfseries 0.72 & 0.58 & 0.61 \\
			Precision & \bfseries 0.80 & 0.69 & 0.75\\
			\bottomrule
		\end{tabular}
}	
	\caption{Training on (3DW-real + 3DW-syn) with three sets of pretrained weights: random and obtained by training on (3DW-real) and (3DW-syn), respectively. Testing was done on Test\_B.}
	\label{table:t2}
\end{table}

\section{Conclusion} 
\label{section V}

We showed that domain randomization is an effective technique in the context of watermark retrieval from 3D printed objects. Using synthetic DR data, intentionally disclaiming photorealism to force the neural network focus on the most relevant features, we trained the neural network (CNN-3DW) to generate confidence maps for the locations of the watermark bumps. When synthetic DR data were mixed with real image data the retrieval rates were higher than using real or synthetic data alone.  
%With fine-tuning on real images, we have shown that the synthetic DR data outperforms more only synthetic dataset, and improves upon results using real data alone, when the test image data unseen in training dataset of real images. 

While the process of inferring information about 3D objects from 2D images of them would always be subject to certain inherent limitations, in some applications recourse to the exact CAD models of those 3D objects can facilitate computer vision tasks. In particular, the annotation of the synthetic images of our training dataset is automatic, eliminating the human error and cost of a hand-labelling, and performed in the 3D domain, eliminating a systematic error that is unavoidable in certain situations where information about 3D objects is directly annotated on 2D images. In the future, we plan to test the technique on more general problems of object recognition, object location and counting problems, working with more complex instances of 3D printed objects and their corresponding CAD models. 
	
\bibliographystyle{apalike}
\bibliography{ref}

\begin{thebibliography}{}

\bibitem[Atapour-Abarghouei and Breckon, 2018]{atapour2018real}
Atapour-Abarghouei, A. and Breckon, T.~P. (2018).
\newblock Real-time monocular depth estimation using synthetic data with domain
  adaptation via image style transfer.
\newblock In {\em Proc. CVPR}, volume~18, page~1.

\bibitem[Barron et~al., 1994]{barron1994performance}
Barron, J.~L., Fleet, D.~J., and Beauchemin, S.~S. (1994).
\newblock Performance of optical flow techniques.
\newblock {\em IJCV}, 12(1):43--77.

\bibitem[Bors, 2006]{bors2006watermarking}
Bors, A.~G. (2006).
\newblock Watermarking mesh-based representations of 3-d objects using local
  moments.
\newblock {\em TIP}, 15(3):687--701.

\bibitem[Brown, 1992]{brown1992survey}
Brown, L.~G. (1992).
\newblock A survey of image registration techniques.
\newblock {\em CSUR}, 24(4):325--376.

\bibitem[Dosovitskiy et~al., 2015]{dosovitskiy2015flownet}
Dosovitskiy, A., Fischer, P., Ilg, E., Hausser, P., Hazirbas, C., Golkov, V.,
  Van Der~Smagt, P., Cremers, D., and Brox, T. (2015).
\newblock Flownet: Learning optical flow with convolutional networks.
\newblock In {\em Proc. ICCV}, pages 2758--2766.

\bibitem[Gupta et~al., 2016]{gupta2016synthetic}
Gupta, A., Vedaldi, A., and Zisserman, A. (2016).
\newblock Synthetic data for text localisation in natural images.
\newblock In {\em Proc. CVPR}, pages 2315--2324.

\bibitem[Handa et~al., 2016]{handa2016understanding}
Handa, A., Patraucean, V., Badrinarayanan, V., Stent, S., and Cipolla, R.
  (2016).
\newblock Understanding real world indoor scenes with synthetic data.
\newblock In {\em Proc. CVPR}, pages 4077--4085.

\bibitem[Hou et~al., 2018]{Hou18}
Hou, J.~U., Kim, D.~G., Ahn, W.~H., and Lee, H.~K. (2018).
\newblock Copyright protections of digital content in the age of 3d printer:
  Emerging issues and survey.
\newblock {\em IEEE Access}, 6:44082--44093.

\bibitem[Hou et~al., 2015]{hou20153d}
Hou, J.~U., Kim, D.~G., Choi, S., and Lee, H.~K. (2015).
\newblock 3d print-scan resilient watermarking using a histogram-based circular
  shift coding structure.
\newblock In {\em Proc. IH \& MMSec}, pages 115--121. ACM.

\bibitem[Hou et~al., 2017]{Hou17}
Hou, J.~U., Kim, D.~G., and Lee, H.~K. (2017).
\newblock Blind {3D} mesh watermarking for {3D} printed model by analyzing
  layering artifact.
\newblock {\em IEEE Tr. on Inf. Forensics and Security}, 12(11):2712--2725.

\bibitem[Jaderberg et~al., 2014]{jaderberg2014synthetic}
Jaderberg, M., Simonyan, K., Vedaldi, A., and Zisserman, A. (2014).
\newblock Synthetic data and artificial neural networks for natural scene text
  recognition.
\newblock {\em arXiv preprint arXiv:1406.2227}.

\bibitem[Kingma and Ba, 2014]{kingma2014adam}
Kingma, D.~P. and Ba, J. (2014).
\newblock Adam: A method for stochastic optimization.
\newblock {\em arXiv preprint arXiv:1412.6980}.

\bibitem[Krizhevsky et~al., 2012]{krizhevsky2012imagenet}
Krizhevsky, A., Sutskever, I., and Hinton, G.~E. (2012).
\newblock Imagenet classification with deep convolutional neural networks.
\newblock In {\em NIPS}, pages 1097--1105.

\bibitem[Luo and Bors, 2011]{luo2011surface}
Luo, M. and Bors, A.~G. (2011).
\newblock Surface-preserving robust watermarking of 3-d shapes.
\newblock {\em TIP}, 20(10):2813--2826.

\bibitem[Macq et~al., 2015]{macq2015applicability}
Macq, B., Alface, P.~R., and Montanola, M. (2015).
\newblock Applicability of watermarking for intellectual property rights
  protection in a 3d printing scenario.
\newblock In {\em Proc. Web3D}, pages 89--95. ACM.

\bibitem[Mundhenk et~al., 2016]{mundhenk2016large}
Mundhenk, T.~N., Konjevod, G., Sakla, W.~A., and Boakye, K. (2016).
\newblock A large contextual dataset for classification, detection and counting
  of cars with deep learning.
\newblock In {\em ECCV}, pages 785--800. Springer.

\bibitem[Ohbuchi et~al., 2002]{ohbuchi2002frequency}
Ohbuchi, R., Mukaiyama, A., and Takahashi, S. (2002).
\newblock A frequency-domain approach to watermarking 3d shapes.
\newblock {\em Computer Graphics Forum}, 21(3):373--382.

\bibitem[Ojala et~al., 2000]{ojala2000gray}
Ojala, T., Pietik{\"a}inen, M., and M{\"a}enp{\"a}{\"a}, T. (2000).
\newblock Gray scale and rotation invariant texture classification with local
  binary patterns.
\newblock In {\em ECCV}, pages 404--420. Springer.

\bibitem[Otsu, 1979]{otsu1979threshold}
Otsu, N. (1979).
\newblock A threshold selection method from gray-level histograms.
\newblock {\em IEEE transactions on systems, man, and cybernetics},
  9(1):62--66.

\bibitem[{Pov-Ray}, 2018]{povray}
{Pov-Ray} (2018).
\newblock {POV-Ray}persistence of vision raytracer.
\newblock \url{http://www.povray.org/}.
\newblock Accessed on 07.10.2018.

\bibitem[Qiu and Yuille, 2016]{qiu2016unrealcv}
Qiu, W. and Yuille, A. (2016).
\newblock Unrealcv: Connecting computer vision to unreal engine.
\newblock In {\em Proc. ECCV}, pages 909--916. Springer.

\bibitem[Simonyan and Zisserman, 2014]{simonyan2014very}
Simonyan, K. and Zisserman, A. (2014).
\newblock Very deep convolutional networks for large-scale image recognition.
\newblock {\em arXiv preprint arXiv:1409.1556}.

\bibitem[Szegedy et~al., 2015]{szegedy2015going}
Szegedy, C., Liu, W., Jia, Y., Sermanet, P., Reed, S., Anguelov, D., Erhan, D.,
  Vanhoucke, V., and Rabinovich, A. (2015).
\newblock Going deeper with convolutions.
\newblock In {\em Proc. CVPR}, pages 1--9.

\bibitem[Tobin et~al., 2017]{tobin2017domain}
Tobin, J., Fong, R., Ray, A., Schneider, J., Zaremba, W., and Abbeel, P.
  (2017).
\newblock Domain randomization for transferring deep neural networks from
  simulation to the real world.
\newblock In {\em IROS}, pages 23--30. IEEE.

\bibitem[Tremblay et~al., 2018]{tremblay2018training}
Tremblay, J., Prakash, A., Acuna, D., Brophy, M., Jampani, V., Anil, C., To,
  T., Cameracci, E., Boochoon, S., and Birchfield, S. (2018).
\newblock Training deep networks with synthetic data: Bridging the reality gap
  by domain randomization.
\newblock In {\em Proc. CVPRW}, pages 969--977.

\bibitem[Yosinski et~al., 2014]{yosinski2014transferable}
Yosinski, J., Clune, J., Bengio, Y., and Lipson, H. (2014).
\newblock How transferable are features in deep neural networks?
\newblock In {\em NIPS}, pages 3320--3328.

\bibitem[Zhang et~al., 2018]{zhang2018watermark}
Zhang, X., Wang, Q., Breckon, T., and Ivrissimtzis, I. (2018).
\newblock Watermark retrieval from 3d printed objects via convolutional neural
  networks.
\newblock {\em arXiv preprint arXiv:1811.07640}.

\bibitem[Zhang et~al., 2016a]{zhang2016unrealstereo}
Zhang, Y., Qiu, W., Chen, Q., Hu, X., and Yuille, A. (2016a).
\newblock Unrealstereo: A synthetic dataset for analyzing stereo vision.
\newblock {\em arXiv preprint arXiv:1612.04647}.

\bibitem[Zhang et~al., 2016b]{zhang2016single}
Zhang, Y., Zhou, D., Chen, S., Gao, S., and Ma, Y. (2016b).
\newblock Single-image crowd counting via multi-column convolutional neural
  network.
\newblock In {\em Proc. CVPR}, pages 589--597.

\end{thebibliography}
\end{document}